\documentclass[letterpaper, 10 pt, conference]{ieeeconf}  
\IEEEoverridecommandlockouts
\usepackage[colorlinks=true, linkcolor=blue, citecolor=blue, urlcolor=blue]{hyperref}
\usepackage{threeparttable}
\usepackage{placeins}
\usepackage{subfigure}
\usepackage{amsmath}
\usepackage{cleveref}
\usepackage{multirow}
\usepackage{amsbsy}
\usepackage{graphicx}
\usepackage{float}

\usepackage{dsfont}
\usepackage{breqn}
\usepackage[utf8]{inputenc}
\usepackage{hyperref}
\usepackage{amssymb}

\hypersetup{
    colorlinks=true,  
    linkcolor=blue,   
    urlcolor=blue,    
    citecolor=blue    
}
\IEEEoverridecommandlockouts                              

\title{\LARGE \bf
VGC-RIO: A Tightly Integrated Radar-Inertial Odometry with Spatial Weighted Doppler Velocity and Local Geometric Constrained RCS Histograms}

 \author{Jianguang Xiang, Xiaofeng He, Zizhuo Chen, Lilian Zhang, Xincan Luo, and Jun Mao
\thanks{This research was funded by the National Nature Science Foundation of China, grant number: 62103430, 62103427, and 62073331 and Major Project of Natural Science Foundation of Hunan Province (No. 2021JC0004)}
\thanks{All authors are with the College of Intelligence Science and Technology, National University of Defense Technology, and National Key Laboratory of Equipment State Sensing and Smart Supporty, Changsha 410073, China (e-mail: 1710950393@qq.com; hexiaofeng@nudt.edu.cn; 1316676920@qq.com; lilianzhang@nudt.edu.cn; 394013122@qq.com;  maojun12@nudt.edu.cn).}
\thanks{\textit{Co-first authors: Jianguang Xiang and Xiaofeng He} contribute equally to this work}
\thanks{\textit{Corresponding author: Jun Mao}}
}

\begin{document}

\maketitle
\thispagestyle{empty}
\pagestyle{empty}

\begin{abstract}
 Recent advances in 4D radar-inertial odometry have demonstrated promising potential for autonomous localization in adverse conditions. However, effective handling of sparse and noisy radar measurements remains a critical challenge. In this paper, we propose a radar-inertial odometry with a spatial weighting method that adapts to unevenly distributed points and a novel point-description histogram for challenging point registration. To make full use of the Doppler velocity from different spatial sections, we propose a weighting calculation model. To enhance the point cloud registration performance under challenging scenarios, we construct a novel point histogram descriptor that combines local geometric features and radar cross-section (RCS) features.
We have also conducted extensive experiments on both public and self-constructed datasets. The results demonstrate the precision and robustness of the proposed VGC-RIO.
\end{abstract}

\section{INTRODUCTION} \label{Introduction}

\begin{figure}[t]
	\centering
	\includegraphics[width=1.0\linewidth] {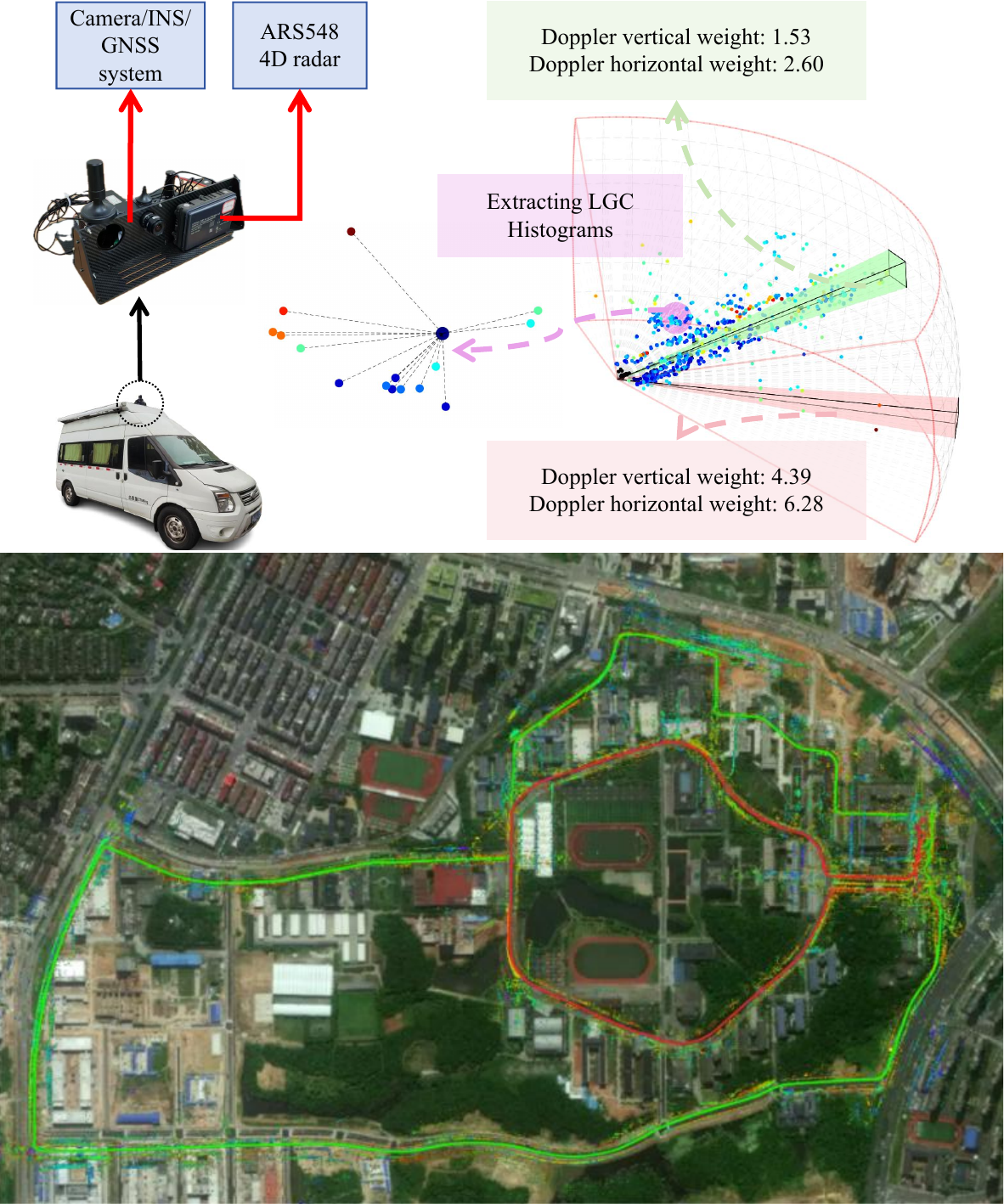}
        \caption{
        Upper left: Vehicle platform and sensors. 
        Upper right: Point cloud segmented into a spherical coordinate system-like structure, with Doppler weights and LGC histograms extracted (details in later sections). Dynamic points shown in black. 
        Lower: Trajectory and mapping results visualized on Google Earth.
    }
	\label{figure:system overview}
\end{figure}

Accurate state estimation is crucial for autonomous vehicles and mobile robots navigating in complex and unknown environments, particularly in areas where GNSS is denied. While LiDAR and camera-based solutions have shown excellent performance, they may fail under adverse weather conditions such as rain, fog, and snow \cite{ref1}. In contrast, millimeter-wave radar, due to its operating frequency band characteristics, remains unaffected by these conditions \cite{ref2, ref3, ref4}, making it an attractive alternative. Particularly, the emerging 4D millimeter-wave radar provides additional elevation information and a denser point cloud than the conventional 3D millimeter-wave radar (hereinafter referred to as 3D radar and 4D radar), thus enabling three-dimensional imaging capabilities similar to those of LiDAR \cite{ref5}. However, 4D radar point clouds remain sparser than LiDAR point clouds and contain more noise and clutter \cite{ref6}.

Current radar registration methods largely follow LiDAR-based approaches, lacking algorithms specifically tailored to the unique characteristics of radar. Meanwhile, while the Doppler information in a single radar frame can be used for velocity estimation, the impact of non-uniform point cloud distribution on velocity estimation has not been adequately addressed.\\
\indent Based on these considerations, we fully leverage the RCS and the spatial distribution information of the point cloud from the radar to propose a tightly integrated radar-inertial odometry method named VGC-RIO. Specifically, we introduce a spatial-distribution-based weighted Doppler residual to enhance the accuracy and robustness of velocity estimation. In addition, we propose a histograms descriptor that combines local geometric features with RCS features to improve point-to-point correspondence registration.
The main contributions of our work are as follows: 
\begin{itemize}
\item Instead of treating the dopper velocity of each point equally, we propose a weighting method to put different weight on each point's vetical and horizental Doppler velocity according to their spatial distribution
\item We propose a novel point-description histogram named LGC, which integrates the spatial distribution of neighboring points and their RCS into a histogram. The LGC histograms helps to register noisy radar points under vigorous motion conditions. 
\item We conducted ablation studies and comparative experiments on diverse and extensive datasets to comprehensively evaluate the performance of our proposed VGC-RIO. Specifically, we evaluated the robustness of the algorithm against sensor jitter on sequences exhibiting different levels of sensor jitter.
\end{itemize}
\hspace{5pt} The remainder of this letter is organized as follows. Section \ref{Related Works} reviews the related work. Section \ref{METHODOLOGY} describes the proposed VGC-RIO method. Section \ref{EXPERIMENTS} presents comprehensive ablation studies on public and self-built datasets, with detailed result analysis. Finally, Section \ref{CONCLUSIONS} concludes the paper and discusses future work.

\section{RELATED WORK} \label{Related Works}
We briefly review the related work on 3D and 4D radar odometry.
\subsection{3D Radar Odometry}
Traditional 3D radars, lacking height information, represent data as 2D point clouds or planar images. Odometry methods can be categorized into feature-based and direct methods.\\
\indent Feature-based methods typically involve keypoint extraction and descriptor matching. Barnes et al. \cite{ref7} proposed a self-supervised framework that uses localization errors to learn keypoint positions, scores, and descriptors. Burnett et al. \cite{ref8} introduced a radar odometry method combining probabilistic trajectory estimation with deep learning features to reduce outliers. Cen et al. \cite{ref9} developed a rotation-invariant descriptor using histograms of angular slices and annular regions, with graph matching for keypoint pairing. Adolfsson et al. \cite{ref10} presented an efficient odometry method for scanning radar, retaining the \textit{k}-strongest echoes to reconstruct surface normals and using point-to-line matching with multi-keyframe registration for accurate localization. Their subsequent work \cite{ref11} addressed sparsity and filtering issues using weighted residuals and joint multi-keyframe registration. Hong et al. \cite{ref12} enhanced pose estimation accuracy and robustness by proposing a novel feature matching method based on radar geometry and graph representation.\\
\indent Direct methods involve processing radar data directly. Park et al. \cite{ref13} proposed a direct radar odometry method that uses the Fourier-Mellin transform to maximize the correlation of log-polar radar images, estimating rotation and translation sequentially to achieve a coarse-to-fine decoupled estimation. Kung et al. \cite{ref14} constructed probabilistic radar submaps to address data sparsity, employing weighted probabilistic normal distributions transform for scan matching. Haggag et al. \cite{ref15} proposed a similar probabilistic approach to estimate the ego-motion using automotive radar.
\subsection{4D Radar Odometry}
Compared with 3D radar, 4D radar provides additional height information, enhanced point cloud quantity and resolution. However, the point cloud density remains low (usually under 1000 points per frame), with significant noise. Current research mainly addresses these issues.\\
\indent In 4D iRIOM \cite{ref16}, a distribution-to-multi-distribution matching approach and a robust GNC are used to handle sparse point clouds and estimate self-velocity from a single scan. 4DRadarSLAM \cite{ref4} introduces an improved APDGICP algorithm that incorporates point probability distribution and RANSAC for velocity estimation, improving localization in low-speed scenarios. Michalczyk et al. \cite{ref17} propose a tightly-coupled Extended Kalman Filter with stochastic cloning for 3D point matching robustness. In their subsequent work \cite{ref18}, persistent surfaces are introduced to enhance localization accuracy by considering stable radar points as landmarks. Li et al. \cite{ref19} develop a 4D radar SLAM framework using pose graph optimization and PCA to extract ground points, removing underground ghost points and improving point cloud quality.\\
\indent Other works enhance registration with RCS or intensity. Huang et al. \cite{ref20} propose a radar-inertial odometry system using RCS for improved point-to-point correspondences. In Milli-RIO \cite{ref21}, intensity information is used to eliminate erroneous matches. Kim et al. \cite{ref22} propose a feature extraction algorithm based on a polar coordinate network utilizing RCS for robust odometry. DGRO \cite{ref23} integrates 4D radar and gyroscope data with RCS for point cloud filtering and weighted registration.\\
\indent Current research on 4D radar odometry primarily focuses on adapting 3D LiDAR odometry methods with minor modifications to fit radar odometry \cite{ref4} \cite{ref16} \cite{ref19} \cite{ref21}, such as the ICP \cite{ref24}, NDT \cite{ref25}, and their variants . Methods based on feature matching, such as LeGO-LOAM \cite{ref26} and Fast-LIO \cite{ref27}, which rely on features like planes and edges, are still challenging due to the sparsity of 4D radar point clouds compared to LiDAR . Additionally, methods using handcrafted feature descriptors, such as FPFH \cite{ref28}, and SHOT \cite{ref29}, are computationally intensive and thus difficult to apply in real-time odometry applications.\\
\indent The VGC-RIO we propose draws inspiration from previous works \cite{ref10}, \cite{ref16}, \cite{ref18}, \cite{ref20}, \cite{ref28}. 

\section{METHODOLOGY} \label{METHODOLOGY}
\begin{figure*}[t]
	\centering
	\includegraphics[width=14cm]{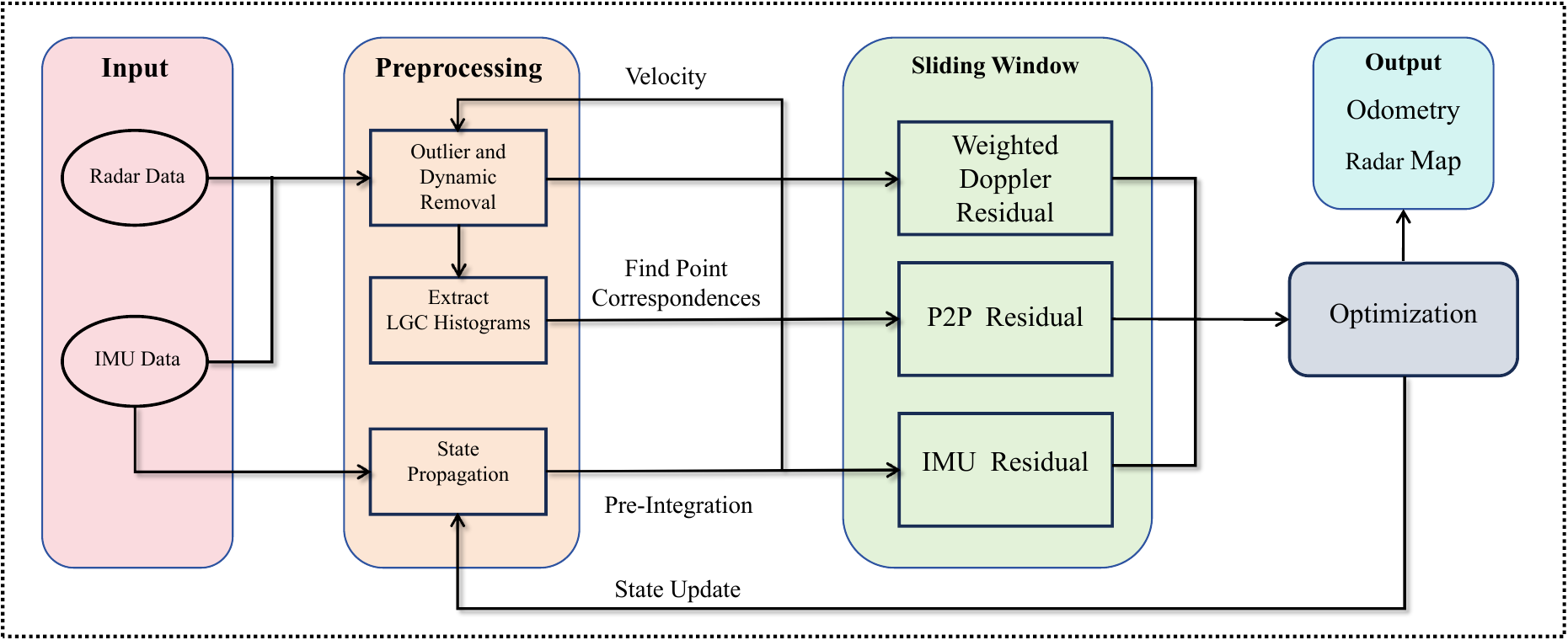}
	\caption{System overview of VGC-RIO}
	\label{figure:pipeline}
\end{figure*}

\subsection{Notations} 
The notations and frame definitions used in this paper are defined in this part. 
The world frame is denoted as $(\cdot)^{w}$.  The body frame, which is consistent with the IMU frame, are denoted as $(\cdot)^{b}$,  while the the radar frame are represented as $(\cdot)^{r}$.
The rotation matrix $\boldsymbol{R}$ and Hamilton quaternions $\boldsymbol{q}$ are used to represent rotations. $\boldsymbol{q}_b^w$ and $\boldsymbol{p}_b^w$ denote the rotation and translation from the body frame to the world frame. $b_{k}$ and  $r_{k}$ respectively represent the body frame and the radar frame at the time when the ${k}$-th frame of radar data is obtained. $\otimes$ denotes the multiplication of two quaternions. $\boldsymbol{g}^w=[0, 0, g]^T$ is the gravity vector in the world frame. Finally, $\hat{(\cdot)}$ indicates estimation of a certain quantity.
\subsection{Problem Formulation}
The proposed VGC-RIO is shown in the Fig. \ref{figure:pipeline}. The system's inputs consist of radar point clouds and 6DoF IMU data. Firstly, preprocessings are implemented to remove dynamic radar points and to divide the points into different sub-regions. 
Then three residual terms are constructed, which are the IMU pre-integration residual, the weighted Doppler residual using static points, and the P2P residual based on LGC histograms for point-to-point correspondences. 
Finally, the three residuals in the sliding window are optimized as a whole with nonlinear solvers. 

The complete state of the system is defined as follows:
\begin{equation}
\begin{aligned}
\mathcal{X} &= [\boldsymbol{\mathit{x}}_{0}, \boldsymbol{\mathit{x}}_{1}, \cdots, {\boldsymbol{x}}_{k}, \boldsymbol{l}{_0^w}, \boldsymbol{l}{_1^w}, \cdots, \boldsymbol{l}{_{m-1}^w}] \\
\boldsymbol{x}_k &= \left[\boldsymbol{p}_{b_k}^w, \boldsymbol{v}_{b_k}^w, \boldsymbol{q}_{b_k}^w, \boldsymbol{b}_{a_k}, \boldsymbol{b}_{g_k}\right]^T, \quad k \in [0, n-1]
\end{aligned}
\label{eq:state}
\end{equation}
\indent where $\boldsymbol{x}_{k}$ represents the IMU state at the time of obtaining the \( k \)-th radar frame. Where $\boldsymbol{p}_{b_k}^w$, $\boldsymbol{v}_{b_k}^w$, $\boldsymbol{q}_{b_k}^w$, $\boldsymbol{b}_{a_k}$ and $\boldsymbol{b}_{g_k}$ denote the position, rotation (represented by a quaternion), velocity, acceleration bias, and gyroscope bias of the \( k \)-th frame, respectively.  $\boldsymbol{l}{_{j}^w} \in \mathds{R}^3$ represents the \( j \)-th landmark.  \( n \) denotes the number of radar frames within the sliding window and \( m \) is the total number of landmarks.
The optimization problem we construct is formulated as follows:

\begin{dmath}
\min_{\mathcal{X}}\left\{
\sum_{k\in I}\left\|r_I\left(\boldsymbol{\hat{z}}_{b_{k+1}}^{b_k}, \mathcal{X}\right)\right\|_{\boldsymbol{P}_k^{k+1}}^2 
+ \sum_{k\in D}\rho\left(\left\|w_{D_k} r_D\left(\boldsymbol{p}_k^r, \boldsymbol{v}_k^r, \mathcal{X}\right)\right\|^2\right)
+ \sum_{k\in D}w_P\rho\left(\left\|r_P\left(\boldsymbol{l}_j^w, \boldsymbol{p}_k^r, \mathcal{X}\right)\right\|^2\right)
\right\}
\label{eq:problem}
\end{dmath}

\indent where $\rho(\cdot)$ is the Huber norm \cite{ref30}. \( w_{D_k} \) and \( w_P \) are residual weights. Detailed explanations of \( w_{D_k} \) and the three types of residuals \( r_I \), \( r_D \), and \( r_P \) are provided in Section \ref{imu pre}, Section \ref{WD}, and Section \ref{p2p}, respectively. 

\subsection{IMU Pre-integration} \label{imu pre}
To avoid the redundancy of repeated integration, we employ the IMU pre-integration. According to \cite{ref31}, the residual of pre-integrated IMU measurements between two consecutive frames \( b_k \) and \( b_{k+1} \) within the sliding window can be defined as:

\begin{equation}
    \begin{aligned}
    & r_I \left( {\hat{\boldsymbol{z}}_{b_{k+1}}^{b_k}}, \mathcal{X} \right) = \begin{bmatrix}
    \delta \boldsymbol\alpha_{b_{k+1}}^{b_k} \\
    \delta \boldsymbol\beta_{b_{k+1}}^{b_k} \\
    \delta \boldsymbol\theta_{b_{k+1}}^{b_k} \\
    \delta \boldsymbol b_a \\
    \delta \boldsymbol b_g
    \end{bmatrix} \\
    &= \begin{bmatrix}
    \boldsymbol{R}_w^{b_k} \left( \boldsymbol{p}_{b_{k+1}}^w - \boldsymbol{p}_{b_k}^w + \frac{1}{2} \boldsymbol{g}^w \Delta t_k^2 - \boldsymbol{v}_{b_k}^w \Delta t_k \right) - \hat{\boldsymbol\alpha}_{b_{k+1}}^{b_k} \\
    \boldsymbol{R}_w^{b_k} \left( \boldsymbol{v}_{b_{k+1}}^w + \boldsymbol{g}^w \Delta t_k - \boldsymbol{v}_{b_k}^w \right) - \hat{\boldsymbol\beta}_{b_{k+1}}^{b_k} \\
    2 \left[ \boldsymbol{q}_{b_k}^{w-1} \otimes \boldsymbol{q}_{b_{k+1}}^w \otimes \left( \hat{\boldsymbol\gamma}_{b_{k+1}}^{b_k} \right)^{-1} \right]_{xyz} \\
    \boldsymbol{b}_{a_{k+1}} - \boldsymbol{b}_{a_k} \\
    \boldsymbol{b}_{g_{k+1}} - \boldsymbol{b}_{g_k}
    \end{bmatrix}
    \end{aligned}
    \label{eq:imu res}
\end{equation}

where \(\hat{\boldsymbol{z}}_{b_{k+1}}^{b_k} = \left[\begin{array}{ccc} \hat{\boldsymbol\alpha}_{b_{k+1}}^{b_k} & \hat{\boldsymbol\beta}_{b_{k+1}}^{b_k} & \hat{\boldsymbol\gamma}_{b_{k+1}}^{b_k} \end{array}\right]\) represents the preintegrated IMU measurements between two radar frames. \(\Delta t_k\) is the time interval between the two radar frames. The operator \([\cdot]_{xyz}\) is used to extract the vector part of a quaternion. Due to space limitations, more details about this formula can be found in \cite{ref31}.

\subsection{Radar Data Preprocessing}
In radar point clouds, the presence of invalid points and dynamic points, if not removed prior to input into the backend, can significantly degrade the accuracy of state estimation. Therefore, preprocessing of the raw radar data is essential. \\
\indent \textbf{\textit{1) Outlier Removal: }} 
We adopt the methods used in \cite{ref19} to filter out invalid points. First, based on the IMU data between the two frames and the velocity and angular velocity provided by the odometry from the previous frame, the transformation between \(\boldsymbol{r}_k\) and \(\boldsymbol{r}_{k+1}\) is calculated. Then, we construct a kd-tree for \(\boldsymbol{P}_{k+1}^{r_{k+1}}\) and perform a fixed-radius nearest neighbor search. If there are no neighboring points from \(\boldsymbol{P}_{k}^{r_{k+1}}\) within a radius of \(\delta_p\), the point is marked as a random point and removed.\\
\indent \textbf{\textit{2) Dynamic Removal:}}
Similar to the method in \cite{ref20}, we use the IMU to assist in filtering out dynamic points. Specifically, we use the estimated velocity \(\hat{\boldsymbol{v}}_{b_k}^w\) from the previous frame and the IMU data obtain the initial estimate of the velocity \(\hat{\boldsymbol{v}}_{b_{k+1}}^w\) for the current frame. If the radar point \(\boldsymbol{p}_i^{r_{k+1}}\) in the current frame is a static point in the environment, it should satisfy:
\begin{equation}
\boldsymbol{v}_{error_i}=\left( \frac{\boldsymbol{p}_i^{r_{k+1}}}{\|\boldsymbol{p}_i^{r_{k+1}}\|} \right)^T \boldsymbol{R}_b^r \hat{\boldsymbol{R}}_w^{b_{k+1}} \hat{\boldsymbol{v}}_{b_{k+1}}^w + {\boldsymbol{v}}_i^{r_{k+1}} \approx 0
\label{eq:Doppler error}
\end{equation}
where, \({\boldsymbol{R}}_b^r\) is the rotation matrix from the body frame to the radar frame, and \(\hat{\boldsymbol{R}}_w^{b_{k+1}}\) is the estimated rotation matrix from the world frame to the current body frame. \({\boldsymbol{v}}_i^{r_{k+1}}\) is the Doppler velocity measured for the point \({\boldsymbol{p}}_i^{r_{k+1}}\) (The Doppler velocity in radar is a radial velocity and is a scalar quantity, with negative values indicating that the target is moving towards the radar). 
Firstly, we preliminarily remove points with large velocity errors, as shown below:
\begin{equation}
\mathcal{P}_{\text{static}}^{r_{k+1}} = \left\{ \boldsymbol p_i^{r_{k+1}} \mid |\boldsymbol v_{\text{error}_i}| < v_{\text{thr}} \text{ and } \left| \frac{\boldsymbol v_{\text{error}_i}}{\boldsymbol v_i^{r_{k+1}}} \right| < p_{\text{thr}} \right\}
\end{equation}
where, \(\mathcal{P}_{\text{static}}^{r_{k+1}}\) represents the set of static points in the current frame. \(|\cdot|\) denotes the Euclidean norm. \(v_{\text{thr}}\) and \(p_{\text{thr}}\) are the velocity error threshold and the percentage threshold of velocity error, respectively. \\
\indent \textbf{\textit{3) Cloud Division: }} 
In order to extract key points and count the number of points in different directional intervals subsequently, we divided the entire point cloud into different intervals. Concretely, we divide the point cloud into several equal-length intervals based on the azimuth and elevation. Taking the azimuth as an example, it is shown as follows:
\begin{equation}
\mathcal{P}_k = \left\{p_i \mid k \leq \frac{\theta_i - \theta_{start}}{\theta_{res}} < k+1\right\}, \quad k \in [0, s-1]
\end{equation}
where, \(\mathcal{P}_k\) represents the set of points belonging to the \(k\)-th interval. \(s\) is the number of intervals. \(\theta_{\textit{start}}\) is the starting azimuth, \(\theta_{\textit{res}}\) is the interval spacing, and \(\theta_i\) is the azimuth of \(p_i\). The aforementioned operation is analogous to establishing a discrete spherical coordinate system, which involves only two angular parameters and does not pertain to the radial distance. As shown in Fig. \ref{figure:system overview}.
The purpose of this approach will be demonstrated in Sections \ref{WD} and Sections \ref{LGC}.

\subsection{Weighted Doppler Velocity Residual} \label{WD}
\begin{figure}
    \centering
    \includegraphics[width=0.8\linewidth]{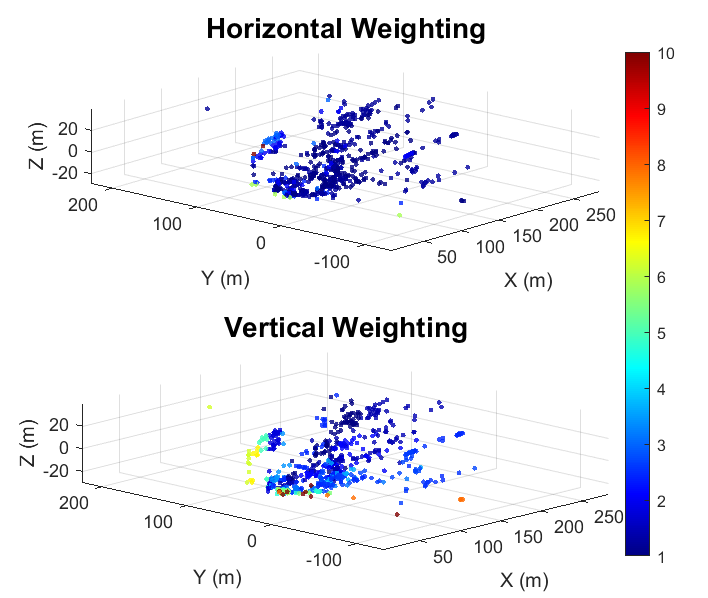}
    \caption{Visualization of weights at different points and in different directions. The point cloud is colored according to the weights.}
    \label{figure:weighted Doppler}
\end{figure}

Ideally, the Doppler velocity measured by radar should correspond to the projection of the target-radar relative velocity vector along the radar's line-of-sight direction. In the data preprocessing step, the dynamic points have already been filtered out. Therefore, for the retained static points, they should satisfy Eq. \ref{eq:Doppler error}. The unweighted Doppler velocity residual is defined as:
\begin{equation}
r_D (\boldsymbol{p}_k^r, \boldsymbol{v}_k^r, \mathcal{X}) = \left( \frac{\boldsymbol{p}_k^r}{\left\|\boldsymbol{p}_k^r\right\|} \right)^T \boldsymbol{R}_b^{r} \hat{\boldsymbol{R}}_w^{b_k} \hat{\boldsymbol{v}}_{b_k}^w + \boldsymbol{v}_i^{r_k}
\label{eq:d res}
\end{equation}
\indent Unlike LiDAR, radar provides sparse point clouds and tends to concentrate reflection points. Therefore, treating all points equally\cite{ref17} \cite{ref18} \cite{ref19} \cite{ref20} may result in an excessive concentration of constraints in directions with dense point clouds, while neglecting constraints in directions with sparse point clouds. This is because points in the same direction provide similar constraints for velocity estimation. To make better use of all points in the scan, we propose weighting the points according to their spatial distribution.
Based on the point cloud division in the preprocessing, \( s \) azimuth intervals \(\left\{ \Omega_k \right\}_{k=1}^{s}\) and \( t \) elevation intervals \(\left\{ \Theta_k \right\}_{k=1}^{t}\) are obtained. Specifically, considering the azimuth intervals as an example, the total constraint of the \(k\)-th azimuth interval is:
\begin{equation}
    C_{k} = \sum_{i \in \Omega_{k}} (w_{k} \cdot r_{i})^{2}
\end{equation}
where $r_{i}$ denotes the Doppler residual of the $i$-th point. Under the assumption that all static points have approximately equal Doppler residuals ($r_i \approx \overline{r}$), the above equation can be expressed as:
\begin{equation}
    C_{k} = \sum_{i \in \Omega_{k}} (w_{k} \cdot \overline{r})^{2}
\end{equation}
\indent The guiding principle of weight calculation is that the constraints in all directional intervals should be kept the same. Then it satisfies:
\begin{equation}
    w_{k}^{2} \cdot n_{k} \cdot \vec{r}^{2} = C
\end{equation}
where \( C \) is a constant, and \( n_{k} \) is the number of points in the \( k \)-th directional interval. By eliminating the influence of the constant term and \( \overline{r} \), the Doppler weight for the \( k \)-th azimuth interval is:
\begin{equation}
w_k = \frac{1}{\sqrt{n_k}}
\end{equation}
\indent The visualization of the weights is shown in Fig. \ref{figure:weighted Doppler}. The weighted Doppler residual are then defined as follows:
\begin{equation}
\begin{split}
w_{D_k} r_D (\boldsymbol{p}_k^r, \boldsymbol{v}_k^r, \mathcal{X}) = 
& \begin{bmatrix}
w_{D_{k_i}} \sin \theta \\
w_{D_{k_j}} \cos \theta
\end{bmatrix}
\cdot \\
& \biggl( \left( \frac{\boldsymbol{p}_k^r}{\left\|\boldsymbol{p}_k^r\right\|} \right)^T \boldsymbol{R}_b^{r} \hat{\boldsymbol{R}}_w^{b_k} \hat{\boldsymbol{v}}_{b_k}^w  + \boldsymbol{v}_i^{r_k} \biggr)
\end{split}
\label{eq:wd res}
\end{equation}
where
\begin{equation}
\begin{cases}
w_{D_{k_i}} = \frac{1}{\sqrt{n_i}} &  i \in [0, s-1] \\
w_{D_{k_j}} = \frac{1}{\sqrt{n_j}} &  j \in [0, t-1]
\end{cases}
\end{equation}
where \(i\) and $n_{i}$ represent the sequence number and the number of points in the azimuth interval to which $\boldsymbol{p}_{k}^{r}$ belongs, respectively. \(j\) and $n_{j}$ represent the sequence number and the number of points in the elevation interval to which $\boldsymbol{p}_{k}^{r}$ belongs, respectively. \(\theta\) represents the elevation.\\
\indent Weights are assigned lower for dense directions and higher for sparse directions to achieve more comprehensive constraints. To enhance the robustness of the estimation, the weights are normalized to the interval [1, 10].

\subsection{Local Geometric constrained RCS histograms} \label{LGC}
As mentioned earlier, a single feature alone is typically not sufficient to provide a comprehensive description. Therefore, we propose a feature histograms descriptor that combines local geometric feature with the RCS feature. 
The detail of RCS can be found in \cite{ref32}. Here, we directly present the assumptions followed by the proposed method: (1) The RCS remains nearly invariant between two consecutive radar scan frames. (2) Points with higher RCS are usually stable and persistent.\\
\indent Subsequently, we provide a detailed exposition on the process of histogram construction and matching: \\
\indent \textbf{\textit{1) Histogram Construction:}}
In the radar data preprocessing step, we obtain several intervals through cloud division. We then extend the \textit{K}-strong filtering algorithm, originally proposed in \cite{ref10}, to a three-dimensional approach. To elaborate, we identify the \textit{k} points (\textit{k} is set to 30 in the subsequent experiments) with the highest RCS values in each interval as key points, which collectively form the key point cloud \( P_{\text{key}} \). This processing ensures that the retained points are spatially uniformly distributed and guarantees their stable existence (based on assumption (2)). Subsequently, we extract the nearest neighbor distance and RCS features for each point in \( P_{\text{key}} \) and encode them into a 2D histogram. Specifically, we first construct a kd tree for \( P_{\text{key}} \), and then traverse each point \( p_i \) in \( P_{\text{key}} \) (\( i \in [0, l-1] \)). For each point, we find its \( k \) nearest neighbors \( p_{n_j} \) (\( j \in [0, k-1] \)). 
The distance and RCS features are defined as follows: 
\begin{equation}
f_{dis} = \left\|{p}_i -{p}_{n_j} \right\|
\end{equation}
\begin{equation}
f_{rcs} =  \mathrm{RCS}_{n_j}
\end{equation}
where, $\left\| \cdot \right\|$ represents the Euclidean distance, and $\mathrm{RCS}_{n_j}$ represents the RCS of \(p_{n_j}\).\\
\begin{figure} [ht!]
    \centering
    \includegraphics[width=1\linewidth]{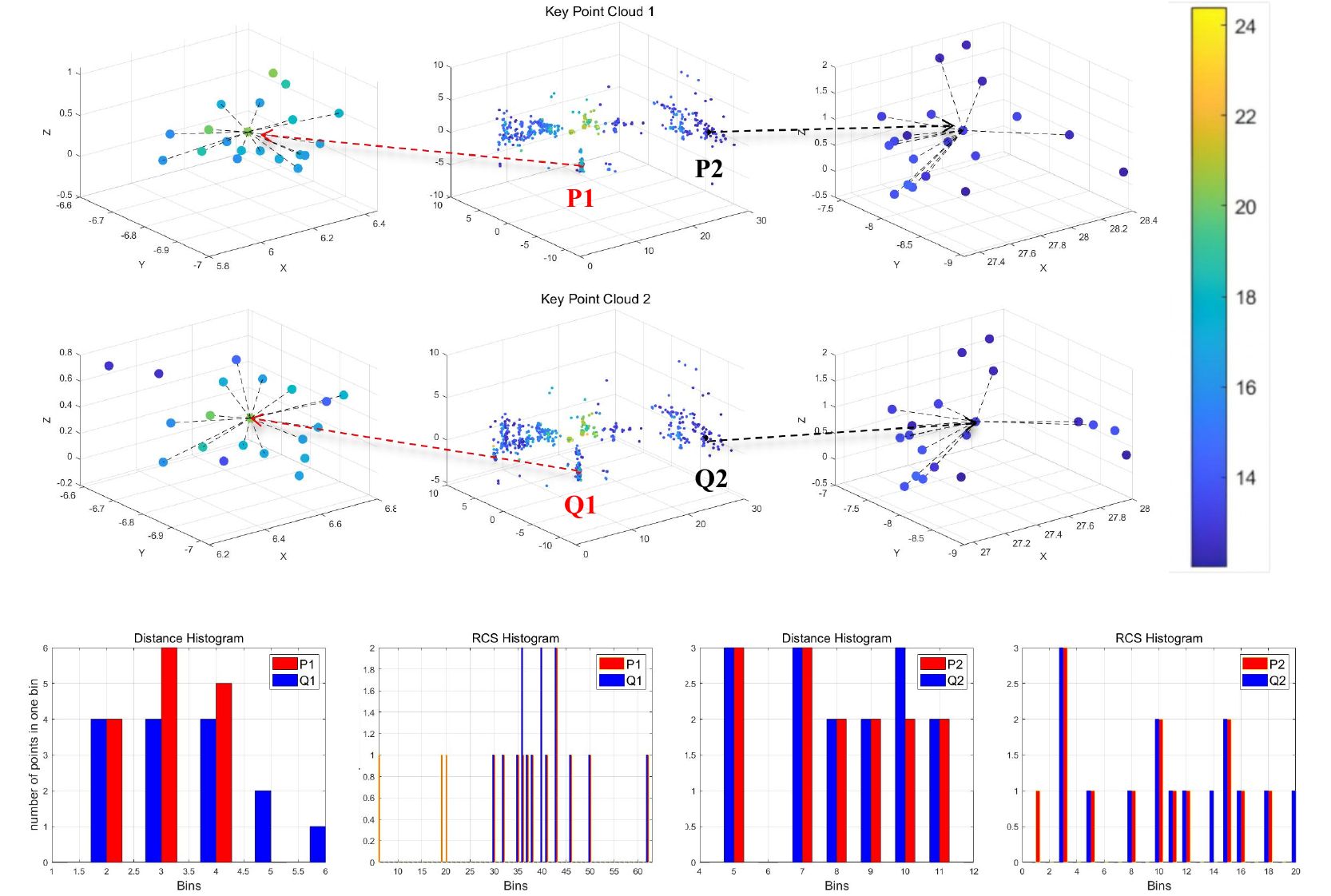}
    \caption{The LGC histograms of corresponding points between consecutive frames. For the sake of visualization, the distance dimension and the RCS dimension of the 2D histogram are compared separately.}
    \label{figure:descriptor}
\end{figure}
\indent The histogram construction for each point \( p_i \) can be represented by the following equation:
\begin{equation}
\begin{aligned}
H_{i}(n,m) &= \sum_{j=0}^{k-1} \mathbb{I}\left(f_{\text{dis}}({p}_i, {p}_{n_j}) \in \text{bin}_n\right) \cdot \\
&\qquad \mathbb{I}\left(f_{\text{rcs}}({p}_{n_j}) \in \text{bin}_m\right), \\
&\quad n \in [0, s-1], \quad m \in [0, t-1]
\end{aligned}
\end{equation}
where \( \mathbb{I}(\cdot) \) is the indicator function that takes the value 1 if the condition is true, and 0 otherwise. Where, \( s \) and \( t \) represent the total number of bins in the distance dimension and the RCS dimension, respectively. The histogram constructed is shown in Fig. \ref{figure:descriptor}.\\
\indent \textbf{\textit{2) Histogram Matching:}} After obtaining the feature histograms of all keypoints in two consecutive frames, we can proceed to identify matching point pairs. To enhance the efficiency of matching, the RCS is employed as an initial screening criterion. Concretely, point pairs with RCS differences exceeding a predefined threshold are deemed non-matching (based on assumption (1)).
\indent We propose Neighborhood-Expanded Histogram Intersection (NHI), a similarity measure for 2D histograms. For histograms \(\boldsymbol{A}\) and \(\boldsymbol{B}\), the NHI similarity \(S_{\!N\!H\!I}\) is defined as:
\begin{equation}
S_{\!N\!H\!I} = \sum_{(i,j) \in \Omega_A} \left[ \max_{\substack{(x,y) \in \Omega_B \\ x \in [i-r,i+r] \\ y \in [j-r,j+r]}} \left( \min(A_{i,j}, B_{x,y}) \cdot W(i,j,x,y) \right) \right]
\end{equation}
where, \( \Omega_A = \{(i, j) \mid A_{i,j} \neq 0\} \), represents the set of coordinates of non-zero elements in histogram \( A \). \( \Omega_B = \{(i, j) \mid B_{i,j} \neq 0\} \), represents the set of coordinates of non-zero elements in histogram \( B \). \( r \) represents the radius of the neighborhood search, which is set to 1 by default. Among them, \( W(i, j, x, y) \) is the distance weight, expressed as:
\begin{equation}
W(i, j, x, y) = \frac{1}{1 + |x - i| + |y - j|}
\end{equation}
where, \(|\cdot|\) denotes the Manhattan distance. If a point pair exhibits the highest similarity and surpasses a predefined threshold, it is considered a successful match. Considering that there may be a few incorrect matches, we employ RANSAC to remove outliers. 
\subsection{Point-to-Point Residual} \label{p2p}
For robust estimation, when constructing residual, we only consider the matched point pairs in the key point cloud and points in the non-key point cloud that have been successfully matched more than a certain number of times (set to 3 in the subsequent experiments). The point-to-point residual are defined as follows:
\begin{equation}
r_P (\boldsymbol{l}_j^w, \boldsymbol{p}_k^r, \mathcal{X}) = \boldsymbol{l}_j^w - \left( \boldsymbol{R}_b^w \left( \hat{R}_r^b 
\boldsymbol{p}_k^r + \boldsymbol{t}_{rb} \right) + \boldsymbol{p}_{b_k}^w \right)
\label{eq:p2p}
\end{equation}
where $\boldsymbol{t}_{rb}$  represents the displacement from \((\cdot)^r\) to \((\cdot)^b\). We use the Ceres-Solver \cite{ref33} to solve the entire optimization problem (Eq. \ref{eq:problem}).
\section{EXPERIMENTS} \label{EXPERIMENTS}
\subsection{Datasets and Setup}
We tested our proposed VGC-RIO system on handheld and vehicle-mounted platforms in the real world, as shown in Fig. \ref{figure:system overview}. The platform is equipped with Continental’s ARS548 4D radar \footnote{https://conti-engineering.com/components/ars-548-rdi/} and a SCHA63T 6DoF IMU \footnote{https://www.murata.com.cn/zh-cn/products/sensor/gyro/overview/lineup/scha63t}. The ground truth was provided by RTK (Real-Time Kinematic) positioning. In addition, we also selected several sequences from the public dataset Snail-Radar \cite{ref34} for testing. \\
\indent We constructed seven sequences, covering a variety of scenarios, as detailed in Tab. \ref{tab:Dataset}. Motion sequences with three different levels of jitter were collected on basketball courts. For court1, we used handheld devices under normal operating conditions, while for court2 and court3, we deliberately shook the devices vertically and horizontally to acquire the data. For the court1, court2, and court3 sequences, the RTK signal is degraded due to occlusion. Given that the starting and ending points of the trajectories coincide, the closure error is employed as the evaluation metric.
\begin{table}[ht]
    \centering
    \renewcommand{\arraystretch}{1.0}
    \begin{threeparttable}
        \caption{Summary of Seven Self-Constructed Datasets} \label{tab:Dataset}
        \begin{tabular}{cccccc}
            \hline
            \textbf{Platform} & \textbf{name} & \textbf{length} & \textbf{speed} & \textbf{shake} & \textbf{dynamic} \\
            \hline
            \multirow{4}{*}{handheld} 
            & court1         & $\approx 300\,\mathrm{m}$         & $\approx 1\,\mathrm{m/s}$         & Mid          & Low \\ 
            & court2         & $\approx 150\,\mathrm{m}$          & $\approx 1\,\mathrm{m/s}$         & High         & Low \\ 
            & court3         & $\approx 200\,\mathrm{m}$          & $\approx 1\,\mathrm{m/s}$         & Extreme         & Low \\
            & garden         & $\approx 300\,\mathrm{m}$          & $\approx 1\,\mathrm{m/s}$         & Mid          & Mid \\  
            \hline
            \multirow{3}{*}{car}
            & loop1    & $2635\,\mathrm{m}$         & $25\,\mathrm{km/h}$       & Low          & Mid \\ 
            & loop2    & $5468\,\mathrm{m}$         & $25\,\mathrm{km/h}$       & Low          & Mid \\     
            & overpass & $7422\,\mathrm{m}$         & $50\,\mathrm{km/h}$       & Mid          & High \\ 
            \hline
        \end{tabular}
        \begin{tablenotes}
            \item[1] “Shake” characterizes the degree of sensor jitter, while “dynamic” indicates the environmental dynamism. Both are categorized into Low, Mid, High, and Extreme levels. 
        \end{tablenotes}
    \end{threeparttable}
\end{table}

\indent Furthermore, we utilized the Snail-Radar dataset to conduct a more comprehensive evaluation of our proposed method. The Snail-Radar dataset comprises two types of 4D radars: the ARS548 and the Oculii EAGLE G7. To verify the generalizability of our algorithm, we performed tests using both types of radars. All experimental results were evaluated using the open-source trajectory evaluation tool EVO \cite{ref35}.\\
\indent The primary parameters of the algorithm are detailed in Tab. \ref{tab:setup}.

\begin{table} [ht!]  
    \centering
    \renewcommand{\arraystretch}{1.0}
    \begin{threeparttable}
    \caption{Algorithm Parameters Setup}

    \begin{tabular}{cccccccccccc}
        \hline
            \multirow{2}{*}{\textbf{parameter}} & \multicolumn{2}{c}{\textbf{Radar Type}} \\
            \cline{2-3} & ARS548 & EAGLE G7 \\
        \hline
             Histogram Distance Bin Width           &0.2   &0.1 \\
             Histogram RCS Bin Width                &1     &0.01 \\
             Histogram Distance Number of Bins      &100   &100 \\
             Histogram RCS Number of Bins           &50    &100 \\
             Neighbor count for histogram           &30    &15 \\
             Threshold of NHI                       &5     &10 \\
        \hline
    \end{tabular}
    \label{tab:setup}
    \end{threeparttable}
\end{table}

\subsection{Ablation Study}
\begin{table}    
\begin{center}
\renewcommand{\arraystretch}{1.0}
\caption{Quantitative Results of Ablation Study on Snail-Radar Datasets}  \label{tab:ABLATION}
\begin{tabular}{cccccc}
\hline
    \textbf{Sequence} &{\textbf{Method}} &{\textbf{APE RMSE (m)}} \\
\hline
    \multirow{4}*{20240113\_1} 
    & D-IMU           & 6.83 \\
    & WD-IMU          & \underline{4.04} \\  
    & P2P-IMU         & 4.32 \\ 
    & Full System     & \textbf{2.50} \\
\hline
    \multirow{4}*{20240116\_eve\_5} 
    & D-IMU           &  27.03 \\ 
    & WD-IMU          &  23.74 \\ 
    & P2P-IMU         &  \underline{17.95} \\ 
    & Full System     & \textbf{15.22} \\ 
\hline
    \multirow{4}*{20231213\_3} 
    & D-IMU         & 109.14 \\
    & WD-IMU        & \underline{69.30} \\  
    & P2P-IMU       & Failed \\ 
    & Full System   & \textbf{67.27} \\ 
\hline
    \multirow{4}*{20240116\_eve\_3}
    & D-IMU         & 130.81\\ 
    & WD-IMU        & \underline{50.17}\\
    & P2P-IMU       & 87.10\\ 
    & Full System   & \textbf{47.26}\\ 
\hline
\end{tabular}
\end{center}
\end{table}

To evaluate the contribution of each component to the system, we conducted ablation studies on four sequences of the Snail-Radar dataset using the ARS548 4D radar. We evaluated four systems: 
\begin{itemize}
\item D-IMU: Unweighted Doppler residuals (Eq. \ref{eq:d res}) with IMU residuals (Eq. \ref{eq:imu res}).
\item WD-IMU: Weighted Doppler residuals (Eq. \ref{eq:wd res}) with IMU residuals.
\item P2P-IMU: P2P residuals (Eq. \ref{eq:p2p}) with IMU residuals.
\item Full System: Weighted Doppler residuals with P2P residuals with IMU residuals.
 \end{itemize} 
 \begin{figure} [h!]
    \centering
    \subfigure[20240113\_1]{	
		\includegraphics[width=4.0cm]{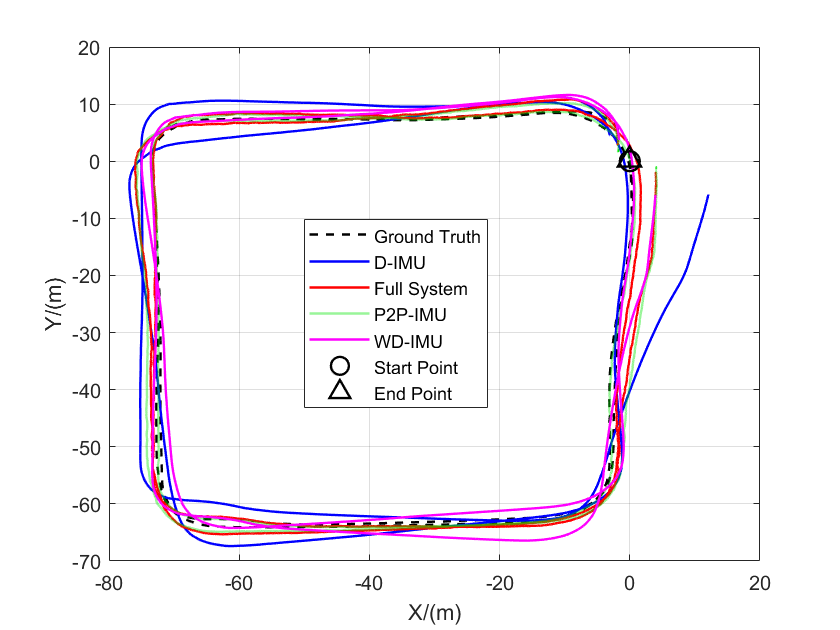}}
    \subfigure[20240116\_eve\_5]{	
		\includegraphics[width=4.0cm]{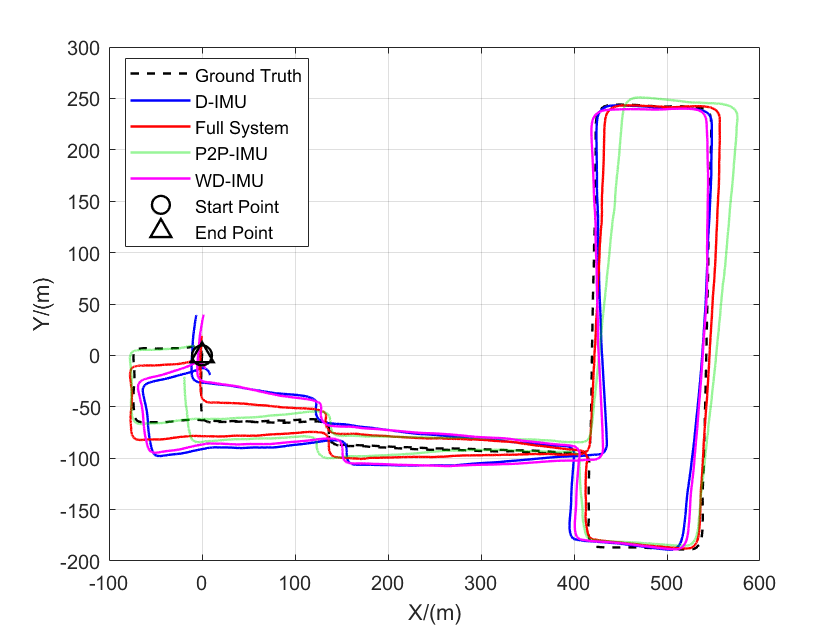}}
    \subfigure[20231213\_3]{	
		\includegraphics[width=4.0cm]{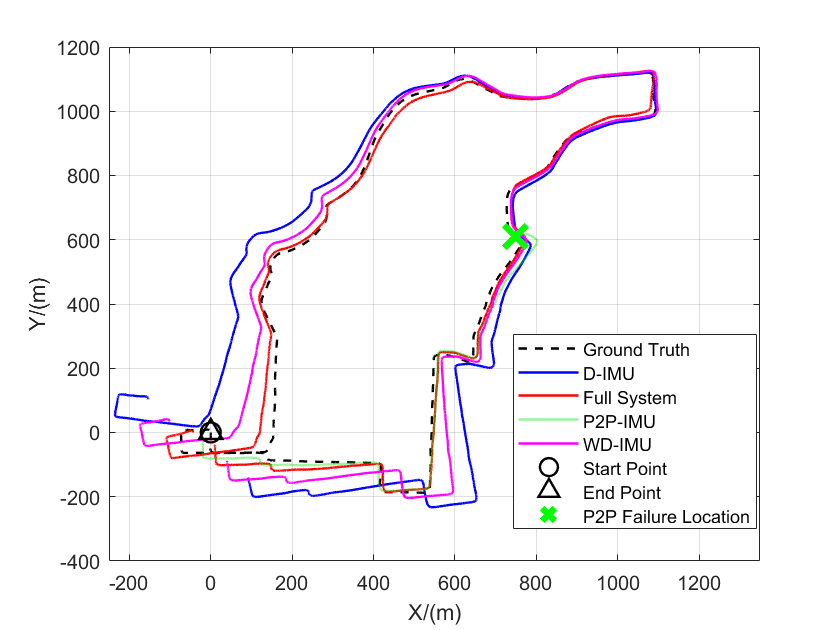}}
  \subfigure[20240116\_eve\_3]{	
		\includegraphics[width=4.0cm]{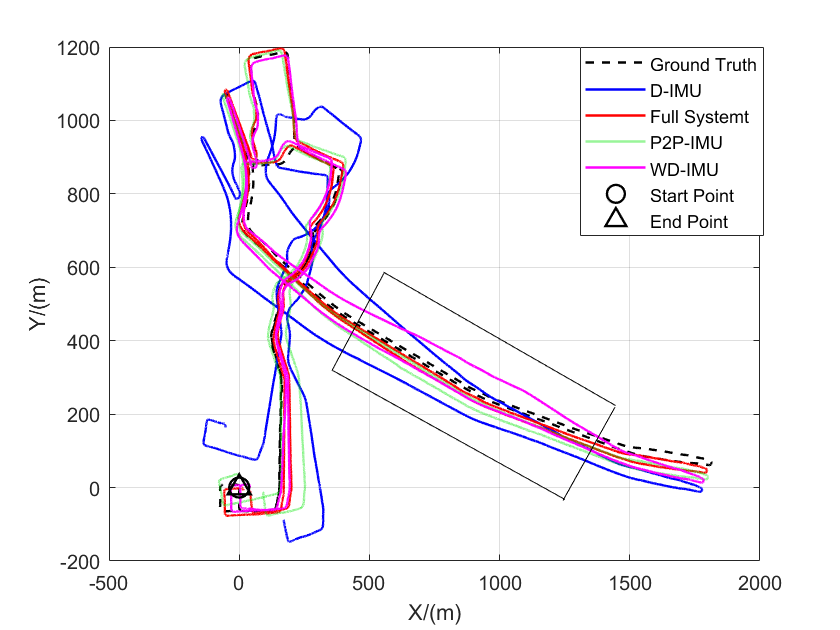}}  
    \caption{The trajectories of four systems on the Snail-Radar dataset}
    \label{figure:ablation traj}
\end{figure}

 \hspace{5pt} The trajectory visualization results of the ablation study are shown in Fig. \ref{figure:ablation traj}. The statistical data of the root-mean-square error (RMSE) are presented in Tab. \ref{tab:ABLATION}. \\
\indent Firstly, the Full System achieved the optimal positioning accuracy across all sequences, which indicates that each component contributes to the system. The combination of the weighted Doppler residuals and the P2P residuals can further enhance the positioning accuracy.\\
\indent In comparison with WD-IMU and D-IMU, it is observed that the positioning accuracy has been improved across all sequences, with particularly significant enhancements in the two long-range sequences (20231213\_3 and 20240116\_eve\_3). This confirms the effectiveness of the weighting strategy.\\
\indent As shown in the black box of Fig. \ref{figure:ablation traj} (d), comparing WD-IMU with the Full System, we find that when the P2P residual constraint is disabled, relying solely on velocity constraints leads to significant drift in the heading angle estimation. This indicates that the P2P residual constraint is crucial for enhancing the stability and accuracy of the heading angle estimation. However, in sparse point cloud scenarios (e.g., 20231213\_3), P2P-IMU failed due to registration difficulties, while WD-IMU operated successfully, further underscoring the value of Doppler velocity measurements.

\subsection{Comparison Experiment}
\begin{figure}[h!]
    \centering

    \subfigure[garden]{	
		\includegraphics[width=4.0cm]{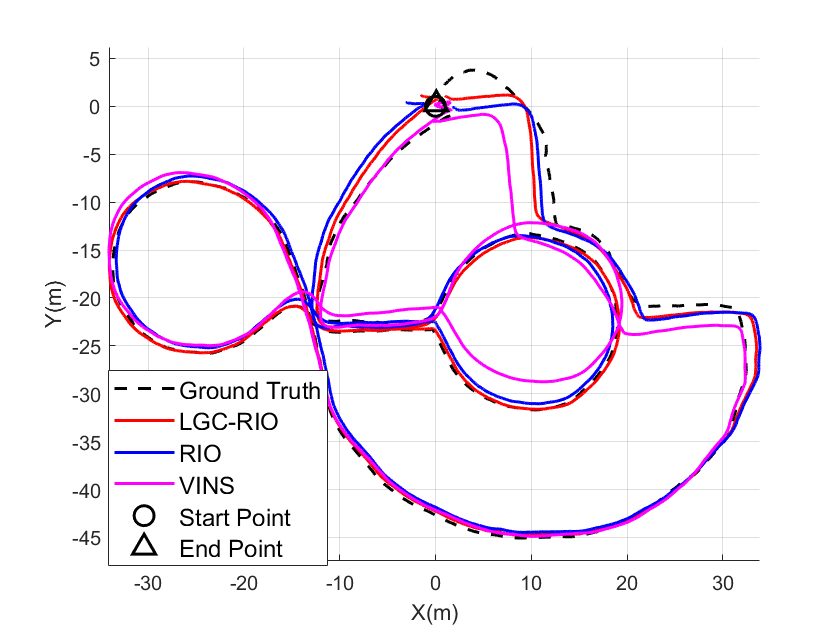}}
    \subfigure[loop1]{	
		\includegraphics[width=4.0cm]{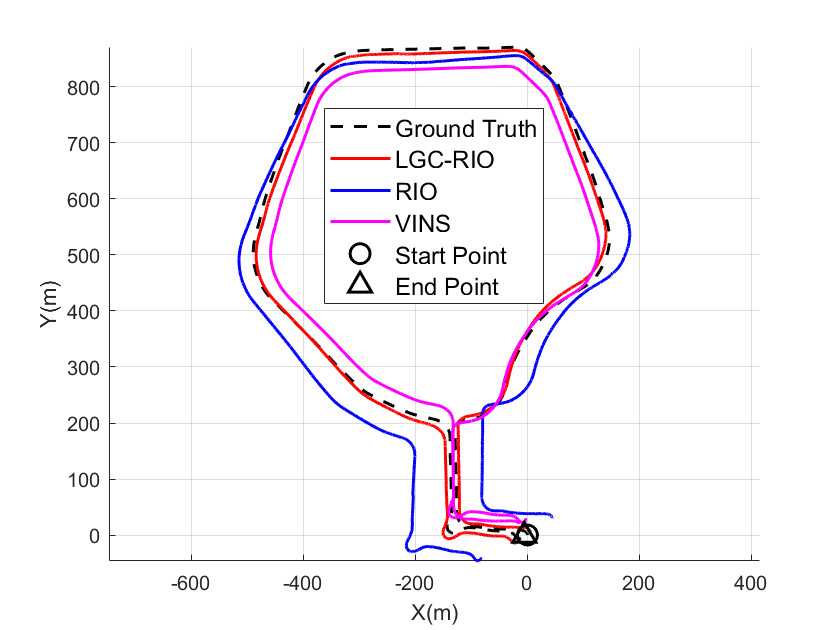}}
    \subfigure[loop2]{	
		\includegraphics[width=4.0cm]{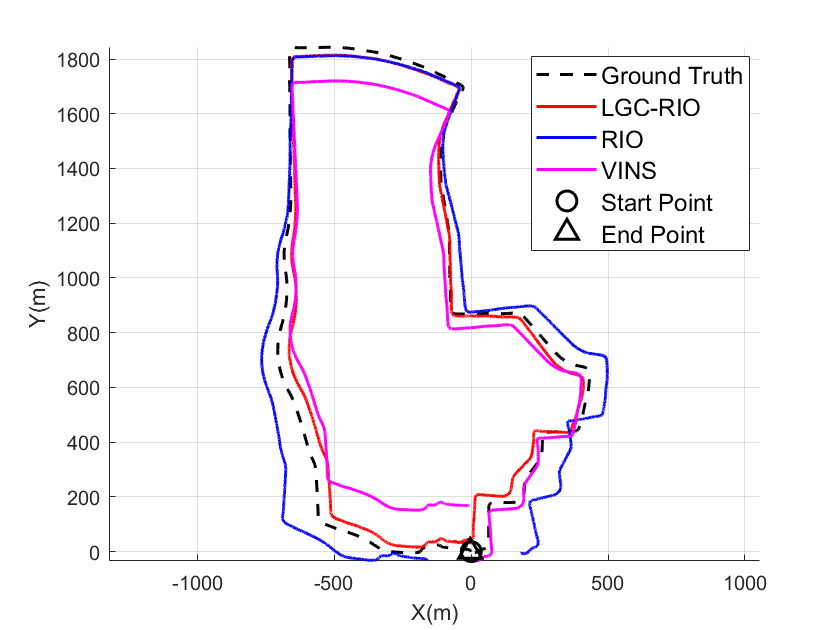}}
  \subfigure[overpass]{	
		\includegraphics[width=4.0cm]{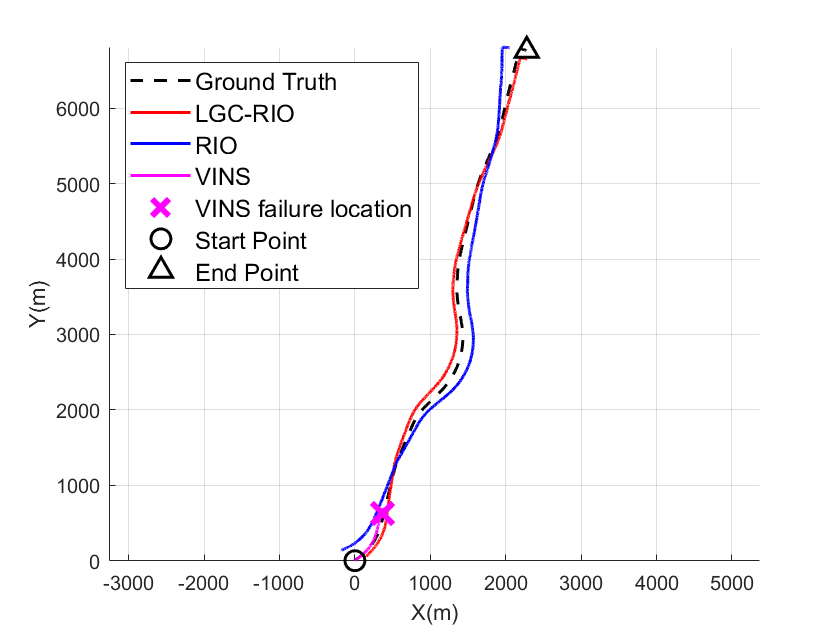}}  
    \caption{Trajectories estimated by the VGC-RIO, RIO and VINS on our four datasets}
    
    \label{figure:compare traj}
    
\end{figure}
We compare the VGC-RIO proposed in this paper with RIO \cite{ref20}, which is considered the current best-performing open-source radar-inertial odometry. In addition, we also selected the classic algorithm VINS \cite{ref31} in visual-inertial odometry for comparison. Since the Snail-Radar dataset also contains the PANDAR XT32 lidar, we further compared it with the lidar-inertial odometry FAST-LIO2 \cite{ref36} on this dataset. Fig. \ref{figure:compare traj} shows the trajectory visualization results of the comparison experiments on our four datasets.Tab. \ref{tab:my compare} and Tab. \ref{tab:snail compare} presents the statistical data of the RMSE of the absolute trajectory errors.\\
\indent First of all, the results presented in Tab. \ref{tab:my compare} indicate that the proposed VGC-RIO achieves the best performance in all sequences and significantly outperforms RIO and VINS in three long-distance sequences. Meanwhile, in the two basketball court sequences with severe jittering, VINS failed to run successfully. RIO succeeded in running on court2 but generated a significant loop closure error, while it failed on court3. In contrast, VGC-RIO was able to run completely in both sequences and maintained good positioning accuracy. This indicates that the proposed VGC-RIO has stronger robustness in aggressive motion scenarios. \\
\indent As indicated by the comparative experimental results presented in Tab. \ref{tab:snail compare}, RIO failed on 20240116\_eve\_3, whereas VGC-RIO successfully processed in all sequences. Moreover, VGC-RIO demonstrated superior localization accuracy across all sequences and different sensors compared with RIO. Additionally, it was found that the localization accuracy of VGC-RIO using the EAGLE G7 radar on the 20231213\_2 sequence even surpassed that of FAST-LIO2.

\begin{table}[ht!]  
    \centering
    \setlength{\tabcolsep}{4pt}
    \renewcommand{\arraystretch}{1.0}
    \begin{threeparttable}
    \caption{Quantitative Analysis: RMSE of Absolute Trajectory Errors (m) on Self-Constructed Datasets}

    \begin{tabular}{c @{\hspace{6pt}} *{7}{c}}
        \hline
            \multirow{2}{*}{\textbf{Method}} & \multicolumn{7}{c}{\textbf{Sequence}} \\
            \cline{2-8} & court1  & court2 & court3 & garden & loop1 & loop2 & overpass\\
        \hline
            VGC-RIO   & \textbf{2.49} & \textbf{3.65}   & \textbf{8.21}   & \textbf{1.79}  & \textbf{15.62} & \textbf{44.01} & \textbf{99.59} \\
            RIO       & \underline{3.36} & \underline{12.33}  & Failed & 6.58  & 53.36  & 110.33 & \underline{124.23}\\
            VINS      & 6.54 & Failed & Failed & \underline{2.04}  & \underline{26.83}  & \underline{95.75}  & Failed \\
        \hline
    \end{tabular}
    \label{tab:my compare}
        \begin{tablenotes}
            \item[1] The sequences of court1, court2, and court3 both have poor RTK signal quality, resulting in the starting and ending points being located at the same position. Therefore, the closure error is employed for evaluation.
        \end{tablenotes}
    \end{threeparttable}
\end{table}

\begin{table}[ht!]  
    \centering
    \renewcommand{\arraystretch}{1.0}
    \begin{threeparttable}
    \caption{Quantitative Analysis: RMSE of Absolute Trajectory Errors (m) on Snail-Radar Datasets}
    
    \begin{tabular}{c @{\hspace{6pt}} c @{\hspace{6pt}} c @{\hspace{6pt}} c} 
        \hline
            \multirow{2}{*}{\textbf{Sensor}} & \multirow{2}{*}{\textbf{Method}} & \multicolumn{2}{c}{\textbf{Sequence}} \\
            \cline{3-4}
            & & \textbf{20231213\_3} & \textbf{20240116\_eye\_3} \\
        \hline
            \multirow{2}{*}{ARS548} & VGC-RIO & 67.27 & \underline{47.26} \\
                  & RIO     & 181.35 & Failed \\
        \hline
            \multirow{2}{*}{EAGLE G7} & VGC-RIO & \textbf{34.67} & 124.09 \\
                     & RIO     & 135.17 & 193.69 \\
        \hline
            XT32 & FAST-LIO2 & \underline{41.23} & \textbf{26.45} \\
        \hline
    \end{tabular}
    \label{tab:snail compare}
    \end{threeparttable}
\end{table}

\section{CONCLUSIONS} \label{CONCLUSIONS}
In this letter, we propose a accurate and robust tightly-coupled 4D radar-inertial odometry system. We introduce a histograms descriptor that combines RCS features and local geometric features. After removing dynamic points, accurate point-to-point correspondences can be obtained through histograms matching. We also propose a spatial-distribution-based weighted Doppler residual, which enables more robust velocity estimation. The proposed VGC-RIO has been validated through ablation studies and comparative analyses on both public datasets and our self-constructed datasets. It demonstrates effectiveness in various environments, including structured and unstructured, low-to-high speed, small-scale and large-scale, as well as Shake Environments. In the future, we will explore the application of radar's RCS feature in loop closure detection.






\end{document}